\pdfoutput=1

\documentclass[11pt]{article}

\usepackage[]{EMNLP2022}

\usepackage{times}
\usepackage{latexsym}
\usepackage{soul}
\usepackage{amsmath}
\usepackage{float, graphicx}

\usepackage{enumitem}

\usepackage[T1]{fontenc}

\usepackage[utf8]{inputenc}

\usepackage{microtype}

\usepackage{inconsolata}

\title{How to disagree well: \\Investigating the dispute tactics used on Wikipedia}

\author{Christine de Kock \\
  Department of Computer \\Science and Technology \\
  University of Cambridge\\
  \texttt{cd700@cam.ac.uk} \\\And
  Tom Stafford \\
 Department of Psychology \\
  University of Sheffield \\
  \texttt{t.stafford@sheffield.ac.uk} \\\And
  Andreas Vlachos \\
  Department of Computer\\Science and Technology \\
  University of Cambridge\\
  \texttt{av308@cam.ac.uk} \\}

\begin{document}
\maketitle
\begin{abstract}
Disagreements are frequently studied from the perspective of either detecting toxicity or analysing argument structure. We propose a framework of dispute tactics which unifies these two perspectives, as well as other dialogue acts which play a role in resolving disputes, such as asking questions and providing clarification. This framework includes a preferential ordering among rebuttal-type tactics, ranging from ad hominem attacks to refuting the central argument. Using this framework, we annotate 213 disagreements (3,865 utterances) from Wikipedia Talk pages. This allows us to investigate research questions around the tactics used in disagreements; for instance, we provide empirical validation of the  approach to
 disagreement recommended by Wikipedia. We develop models for multilabel prediction of dispute tactics in an utterance, achieving the best performance with a transformer-based label powerset model. Adding an auxiliary task to incorporate the ordering of rebuttal tactics further yields a statistically significant increase. Finally, we show that these annotations can be used to provide useful additional signals to improve performance on the task of predicting escalation.  
\end{abstract}

\section{Introduction}
\newcommand\dsname{WikiTactics}
\begin{figure*}
    \centering
    \includegraphics[width=0.95\textwidth]{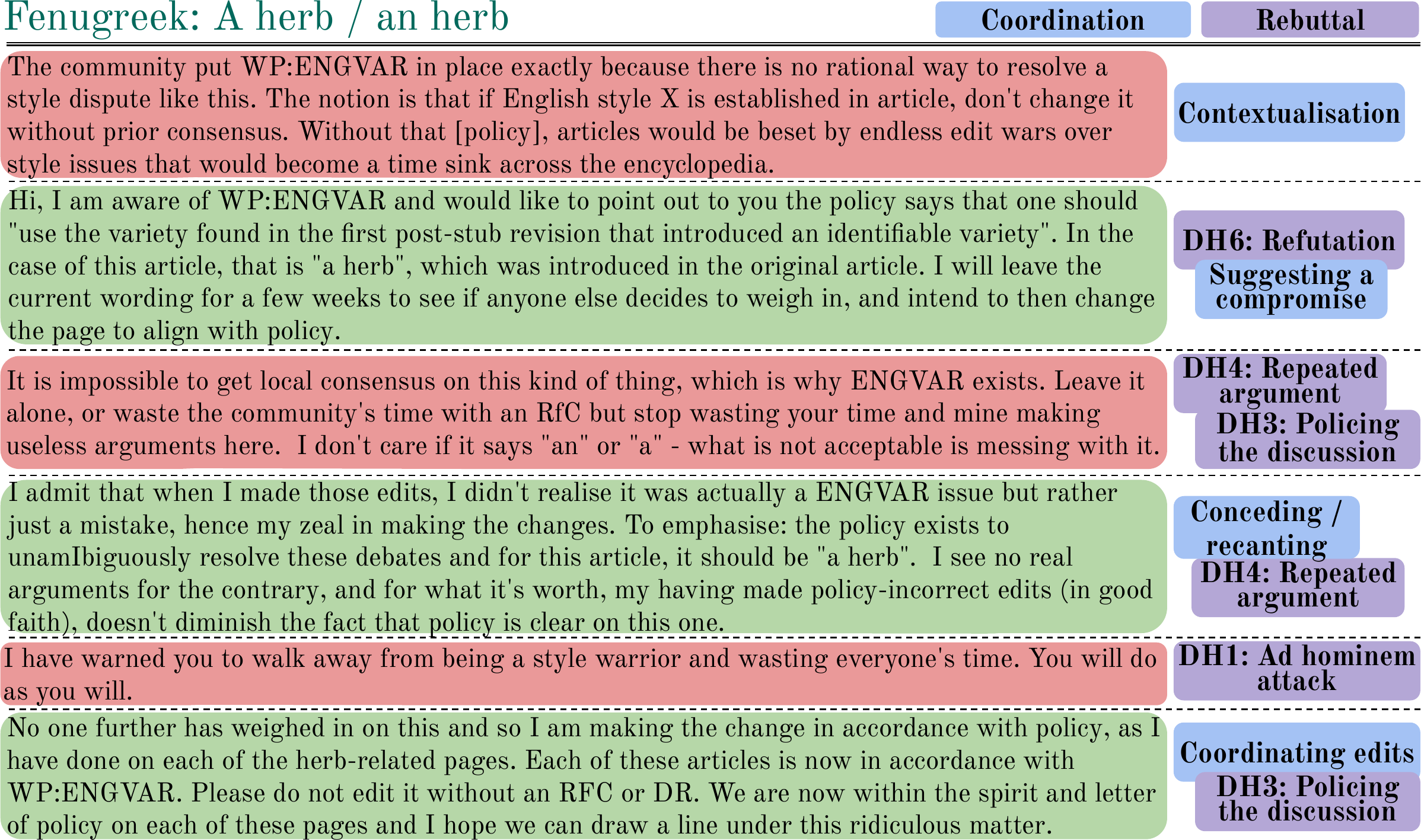}
    \caption{An example from the \dsname{} dataset. Different speakers are indicated by colour. WP:ENGVAR refers to the Wikipedia English Variants policy. RfC and DR refer to Wikipedia dispute resolution procedures.}
    \label{fig:fenugreek}
\end{figure*}
Disagreements are pervasive in online communication. While usually perceived as a negative phenomenon, research in psychology has shown that debate and disagreement can promote beliefs that are better supported by evidence \citep{landemore2012democratic,mercier2018enigma}. \citet{hallsson2020disagreement} argue that, in the ideal case, a range of arguments are considered for both sides, increasing the participants' ability to find good arguments in defence of a view and to critique reasons against it, and thereby forcing one to interact with reasoning and evidence which may otherwise be discarded due to confirmation bias. 

Prior work in NLP has investigated detecting negative artefacts of online disagreements such as personal attacks or hate speech, e.g.\ \citet{wulczyn2017machina} and \citet{waseem-hovy:2016:N16-2}. Research in argumentation mining instead looks at identifying argument structures \cite{lawrence2019argument} and inferring the quality of arguments \citep{habernal-gurevych-2016-argument}, often focusing on classical theories of argumentation (e.g.\ Aristotelian) which do not include less desirable aspects such as personal attacks. However, real world disagreements often contain both well-structured arguments and attacks, in addition to other dialogue acts, such as asking for and providing clarification. 

In this work, we propose a framework of \textit{dispute tactics} consisting of rebuttal and coordination strategies, denoting the role of a particular utterance in the context of a disagreement discussion. We build on the disagreement hierarchy proposed by \citet{graham}, which includes a preferential ordering between different rebuttal tactics.
We introduce \dsname\footnote{\url{github.com/christinedekock11/wikitactics}}: a set of 213 disputes (comprising 3,865 utterances) on Wikipedia Talk pages, manually annotated with the dispute tactics employed in the process of resolving a disagreement between editors. These multiturn, multiparty conversations are sourced from the WikiDisputes dataset \citep{de-kock-vlachos-2021-beg} which is annotated according to whether the dispute was resolved without the need for a moderator. An example of such a conversation is shown in Fig~\ref{fig:fenugreek}. 

Using this framework and data, we investigate a number of research questions related to disagreements. Firstly, we find that a lower mean rebuttal level in a disagreement is correlated with less constructive dispute resolutions, providing empirical validation of the ordering proposed by \citet{graham} and recommended by Wikipedia to its editors. 
Individual users are found to utilise a range of rebuttal levels more often than adhering to only the top or bottom of the rebuttal hierarchy.  
We quantify the use of mirroring in disagreements by observing how users deviate from their own mean rebuttal level depending on the rebuttal level used in a conversation, finding that mirroring takes place in 57\% of cases.

We further develop models for predicting the dispute tactics used in an utterance as a multilabel classification task, experimenting with both the binary relevance and label powerset approaches, as well as models with and without conversation context. Our best model is a transformer-based model which uses a label powerset approach and context. A statistically significant improvement is gained by taking into account the preferential ordering of the rebuttal tactics defined in our scheme using multi-task training. Finally, we illustrate that these annotations can be leveraged to improve performance on the task of predicting whether a dispute will be resolved without escalating to a moderator. 

\section{Online disagreements}\label{sec:background}
Wikipedia Talk pages are a popular source for NLP studies of goal-oriented discussions (e.g.\ \citealp{niculae-danescu-niculescu-mizil-2016-conversational,kittur2016influence}). The Talk pages are linked to specific articles and are used to coordinate edits and in some cases to resolve disputes. WikiDisputes \citep{de-kock-vlachos-2021-beg} is a dataset of Talk page discussions tagged as ``disputes'' by editors. The dataset provides an ``escalation'' label for each dispute: under the Wikipedia dispute resolution policy, disputes which cannot be resolved by editors themselves are escalated to mediation, which is considered to be a proxy for constructiveness.

Wikipedia recommends the hierarchy of disagreement formulated by \citet{graham} as a guide for constructive dispute resolution \citep{wiki_dispres_policy}. Graham's hierarchy posits that there are seven levels of disagreement, ranging from name-calling (at the bottom) to refuting the central point. Descriptions for these levels are provided in Fig~\ref{fig:grahams_hierarchy} in App~\ref{app:dh}. Graham's hierarchy has been used as a framework for research in various fields, including healthcare \citep{cope2015risk}, education \citep{phelps2019decade} and sociology \citep{neven2019make}. Within NLP, \citet{tangline} have used this taxonomy in combination with LDA topic models to distantly annotate and analyse the rationality of online discussions. Despite its popularity, this hierarchy has not been verified empirically.

Individual levels of this hierarchy have been considered in prior work. For instance, \textit{name-calling} and \textit{ad hominem attacks} (levels 0 and 1) can be considered subsets of personal attacks \citep{waseem-hovy:2016:N16-2,chang-danescu-niculescu-mizil-2019-trouble}. \textit{Counterarguments} (level 4) and \textit{refutations} (level 5) are described in terms of arguments and evidence, a topic that has received extensive attention in the field of argumentation mining, which seeks to categorise different elements of an argument as claim or premise (see, e.g.\ \citet{lawrence2019argument}). \textit{Contradiction} (level 3), described as stating the opposing case, is related to the task of stance detection \citep{10.1145/3369026}. Graham’s hierarchy combines these different concepts and proposes that there is an preferential ordering among them which correlates with more favourable disagreement outcomes.

\citet{walker2012corpus} also focus on online disagreements  
and introduced the Internet Arguments Corpus (IAC). This corpus contains utterance-response pairs annotated for the degree of agreement as well as whether the response is emotional versus factual, respectful versus insulting, asking questions vs asserting, and negotiating versus attack. Some of these markers can be mapped to Graham's hierarchy; however, \citet{walker2012corpus} consider pairs of utterances whereas our aim is to understand disagreements as a multiturn conversation. Notably, their taxonomy illustrates that different disagreement markers can be used in combination: an utterance may be insulting and attacking while still being factual. It also captures the fact that within a disagreement, all discussion is not necessarily aimed at countering a proposition by an opponent; negotiating compromises and asking questions plays an important part in how disputes are resolved. 

Another taxonomy that considers concepts related to Graham's hierarchy is that of \citet{benesch2016counterspeech}, on the topic of counterspeech. For instance, ``presentation of facts to correct misstatements or misperceptions'' is similar to \textit{refutation}, the top level of the rebuttal hierarchy. ``Denouncing hateful speech'' is similar to \textit{contradiction}, and ``hostile denouncing'' is similar to \textit{name-calling} and \textit{ad hominem attacks}.  This taxonomy does not provide the explicit arrangement in levels as \citet{graham} does. \citet{graham} further describes disagreements more abstractly, whereas counterspeech can be thought of as a specific form of disagreement. 

Recent work in argumentation has explored the notion of argument quality. For example, \citet{lukin-etal-2017-argument} investigate the effect of personality factors in an audience on the convincingness of an argument. \citet{habernal-gurevych-2016-argument} introduce a corpus of paired arguments on controversial topics, and propose a decision tree for motivating why an argument is more convincing. This includes aspects such as providing an explanation with facts versus attacking the opponent, which aligns with some of the rebuttal levels described above. However, the aforementioned work on argumentation quality does not consider the conversation context but rather looks at monologic arguments.   

\section{\dsname}\label{sec:pilot}
\subsection{Disagreement annotation schema}

\begin{table}[]
    \footnotesize
    \renewcommand{\arraystretch}{1.1}
    \centering
    \begin{tabular}{|p{0.48\textwidth}|}
    \hline
    \textbf{Rebuttal}\\
    \hline
     \textbf{RH0}: Name calling, insults and hostility\\
     \textbf{RH1}: Attacks to credibility of person or argument\\
     \textbf{RH2}: Attempted derailing / off-topic comments*\\
     \textbf{RH3}: Policing the discussion\\
     \textbf{RH4}: Stating your stance\\
     \textbf{RH4}: Repeated argument*\\
     \textbf{RH5}: Counterargument\\
     \textbf{RH6}: Refutation\\
     \textbf{RH7}: Refuting the central point\\
     \hline
     \textbf{Coordination}\\
     \hline
     Coordinating edits\\
     Contextualisation\\
     Asking questions\\
     Providing clarification\\
     Suggesting a compromise\\
     Conceding / recanting\\
     Bailing out\\
     ``I don't know''\\
     \hline
    \end{tabular}
    \caption{Dispute tactics used to annotate \dsname. For rebuttal tactics, the proposed ordering is indicated by level numbers. Levels marked with an asterisk were added to Graham's disagreement hierarchy.} 
    \label{tab:labels}
\end{table}
In this work, we distinguish between two broad categories of dispute tactics: utterances aimed at countering the point of an opponent (which we term \textbf{rebuttal tactics}) and attempts to promote understanding and consensus (referred to as \textbf{coordination tactics}).

For rebuttal tactics, Graham's hierarchy provides a useful basis. We expand the original taxonomy to include categories for ``repeated argument'' and ``attempted derailing or off-topic comments''. The proposed levels and the preferential ordering are shown in Table \ref{tab:labels}, with complete definitions in Table \ref{tab:dh_schema} in App~\ref{app:dh_schema}.

For the coordination tactics, we draw on previous work on Wikipedia Talk page discussions. The taxonomy
of \citet{ferschke2012recognizing} contains the ``information content'' class which encapsulates providing, seeking  or correcting information; similarly, we annotate both \textit{asking questions} and \textit{providing information}. As conversations on this platform are inherently goal-oriented, some discussion is often expended on \textit{coordinating edits} \citep{viegas2007talk,schneider2011understanding}; we include a class to capture this. We further use \textit{contextualisation} to describe the ``opening statement'' of a dispute on Wikipedia). As discussed in Sec~\ref{sec:background}, the IAC also accounts for \textit{negotiating compromises}, which is encouraged by the platform's dispute resolution policy. We also annotate for this category. Finally, we annotate three ``retreat'' moves: saying \textit{I don't know}, \textit{conceding or recanting} of a position, and giving up or \textit{bailing out}. 
Full definitions for these classes are contained in Table \ref{tab:nondh_schema} in App~\ref{app:ndh_schema}. Unlike with the disagreement levels used for rebuttal moves, there is no ordering among these classes. 

Per the work of \citet{walker2012corpus} and \citet{viegas2007talk}, we use multilabel annotation. We allow for up to three rebuttal strategies and two resolution strategies per utterance. 

\subsection{Data annotation}
Using the taxonomy described above, a set of 213 disputes (3,865 utterances) was annotated. The disputes were sampled from WikiDisputes to contain approximately the same number of escalated and non-escalated disputes. The median conversation length is 21 utterances (minimum 5, maximum 44), with an average utterance length of 54 tokens. The average number of speakers is 4. 

Three initial rounds of pilot annotation were carried out to determine the appropriateness of the taxonomy for the data. During each round, five disputes (totalling 194 utterances) were annotated independently by an experienced professional annotator and one of the authors. Following each round, the annotations were compared and definitions expanded to reduce uncertainty. The agreement scores after each round are shown in Table \ref{tab:agreement} in App~\ref{app:agreement}. The Cohen's $\kappa$ improved after each pilot round, as did the Pearson's $r$. The Pearson correlation takes the ordinality of the different levels into account, while this is ignored by the $\kappa$ score.

An initial source of disparity between the annotators was the fact that Graham's taxonomy describes characteristics of individual responses that express disagreement, as opposed to a thread of utterances in which a topic is discussed. For instance, the notion of the ``central point'' (RH6/7) can take on different meanings when a full thread is considered; referring to either the root of the dispute, or the essence of an argument. In the latter case, the central point drifts as various arguments are evaluated, whereas it would remain static in the former case. In our annotation, we opted to define the central point as relative to the utterance to which one is responding (i.e. the central point drifts). 

The main annotation round was executed by the expert annotator, using the definitions set out in App \ref{app:dh_schema} and \ref{app:ndh_schema}. An example from the dataset is shown in Fig~\ref{fig:fenugreek}. The most common label identified is ``counterargument'' (RH5) with 996 uses, followed by ``coordinating edits'' (a coordination tactic) with 972 uses. Approximately a quarter of utterances were assigned more than one label during our annotation. Frequent label combinations are discussed in Sec~\ref{sec:adhomcx}.

\begin{figure*}
    \centering
    \includegraphics[width=\textwidth]{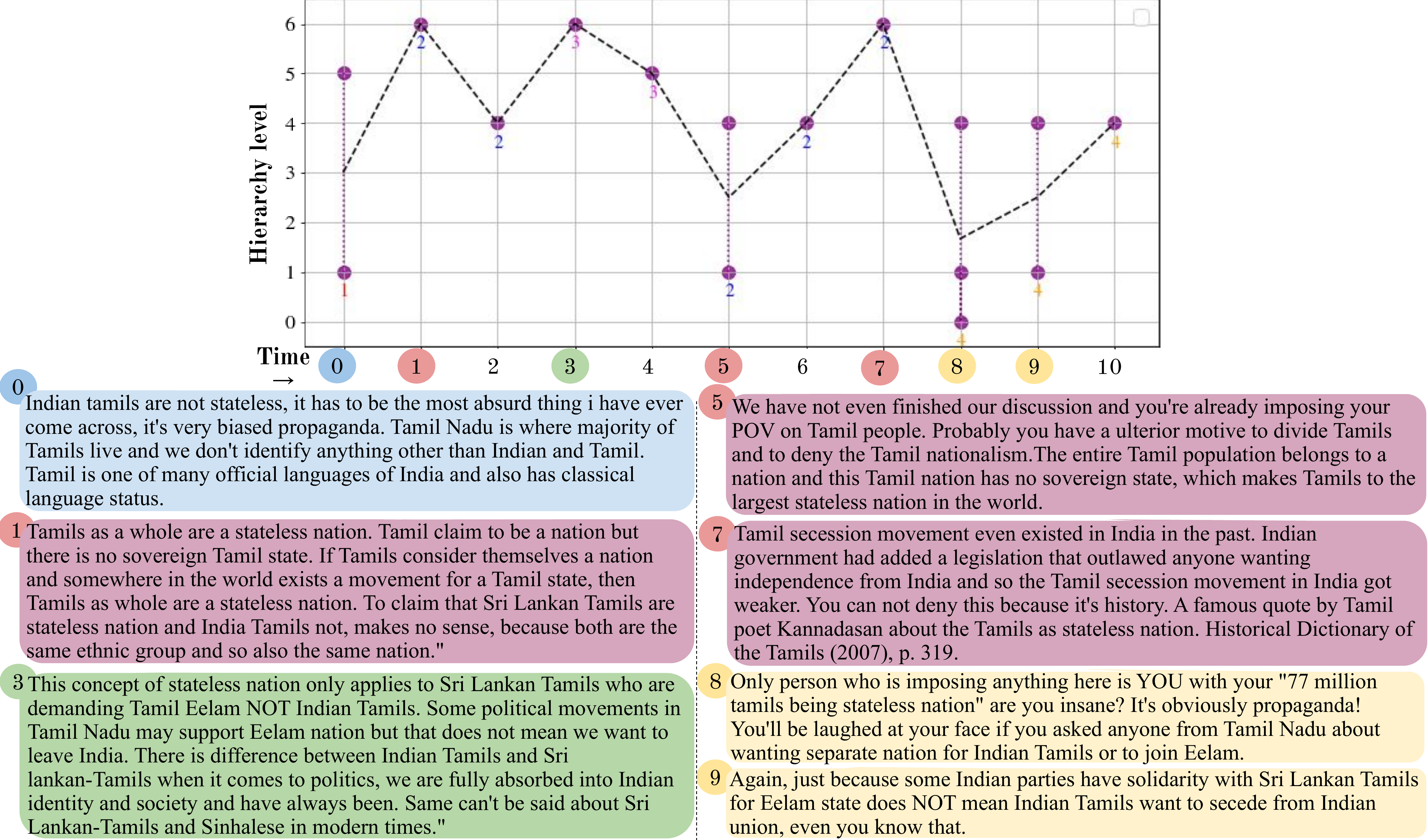}
    \caption{An example of an annotated dispute from \dsname. Different speakers are represented by colours. Some utterances are assigned multiple labels, which are shown as vertical lines.}
    \label{fig:dh_sample}
\end{figure*}

\section{Analysing dispute tactics}\label{sec:analysis}
In this section, we investigate four research questions regarding theories of disagreement:
\begin{enumerate}[noitemsep,nolistsep]
    \item Does the mean rebuttal level in a conversation correlate with escalation?
    \item Which tactics co-occur with personal attacks?
    \item What effect do personal attacks have on a conversation?
    \item Do individual users adhere to certain levels of the rebuttal hierarchy?
\end{enumerate} 

\subsection{Correlation with escalation}\label{sec:correlation}
We are interested in how informative the dispute tactics are in predicting the outcome of a dispute. We consider the escalation tags provided in WikiDisputes and calculate their Spearman correlation with the mean observed rebuttal level in a conversation. We use the micro-averaged mean over all rebuttal scores assigned during the conversation (ie. excluding coordination utterances). Let $C = [({u_1,r_1,c_1}),...,({u_n,r_n,c_n)}]$ denote a  disagreement consisting of utterances $u$, rebuttal labelsets $r$ and coordination labelsets $c$. $l$ represents a mapping of rebuttal labelsets to the numeric values of their levels, as set out in Table \ref{tab:labels} (RH0 - RH7). We then calculate:
\begin{align*}
    s = \frac{\sum_{k=1}^{n}{\sum{l(r_{k})}}}{\sum_{k=1}^{n}{|r_{k}|}}.
\end{align*}
This yields a weak negative correlation ($\rho=-0.19, P=0.005$). We also experimented with a macro-averaged mean, which averages first at the utterance level and then at the disagreement level:
\begin{align*}
    s = \frac{\sum_{k=0}^{n}{\frac{\sum{r_k}}{|r_k|}}}{n}.
\end{align*}
This yielded similar results ($\rho=-0.24, P=0.001$).

Using the mean rebuttal level is an imperfect metric as it ignores temporal order: a dispute starting on a high level but ending with name-calling would have the same mean rebuttal level as the inverse. Nevertheless, this result affirms the intuition that using higher level rebuttals correlates with more constructive outcomes to a disagreement. 

\subsection{The context of personal attacks}\label{sec:adhomcx}
Our annotation scheme defines two types of personal  attacks: \textit{name calling, insults and hostility} (level 0; hereafter referred to as ``name calling'') and \textit{attacks to credibility of the person or the argument} (level 1; hereafter referred to as ``credibility attacks''). We refer to these levels jointly as personal attacks, though it also includes attacking an opponent's argument. Level 1 attacks are the third most common label in the dataset with 575 cases, whereas level 0 attacks occur in only 65 cases. This is aligned with the low levels of toxicity associated with Wikipedia Talk pages \citep{wulczyn2017machina}, but provides the additional insight that members of the Wikipedia community may still be disparaging while avoiding direct insults. 

To investigate the context in which such attacks occur, we evaluate co-occurrences in our multilabel annotation. The most common multilabel combination was \textit{credibility attacks} with \textit{counterargument} (119 occurrences), which is observed more than \textit{credibility attacks} alone. The second most common combination is \textit{credibility attacks} with \textit{repeated argument}. 
To evaluate how much more than chance the classes co-occur, we calculate the pointwise mutual information (PMI; \citealp{jurafskyspeech}) of each label ($x$) with personal attacks ($y$):
\begin{align}
    PMI(x,y) &= log_2(\frac{P(x,y)}{P(x)P(y)}). 
\end{align}

The PMI values for \textit{counterargument} and \textit{repeated argument} with \textit{credibility attacks} are positive, which indicates that they indeed have a larger than chance co-occurrence with attacks. The coordination labels (excluding \textit{bailing out}) have the largest negative PMI values, indicating a less than chance co-occurrence. The full results are shown in Table \ref{tab:pmi} of App~\ref{app:pmi}. 

\subsection{The effects of personal attacks}
Previous work has framed personal attacks as a sign of conversational failure (e.g. \citealp{zhang2018conversations}). Given the above observation that they may co-occur with higher level rebuttals, we are interested in their overall effect on a conversation, and 
therefore calculate the probability of a conversation recovering after a personal attack occurs. We define recovery in terms of having an utterance labeled as rebuttal level 5 or higher and no further personal attacks. By this definition, half of the disputes were found to recover after a personal attack, indicating that personal attacks do not necessarily result in conversational failure. 

Personal attacks are more frequently found in escalated conversations: 60.7\% of personal attacks are in this category (with the total number of escalated and non-escalated disputes being balanced). Of the escalated disputes with personal attacks, only 44.3\% are found to recover, whereas 59.2\% of resolved disputes recover post attack. This indicates that although personal attacks also occur in non-escalated disputes, participants are better adept at moving beyond them. 

We further find that immediate retaliation (i.e.\ a personal attack being followed by another personal attack) occurred in 25.7\% of cases. In disputes where at least one personal attack had occurred, the probability that the initial offender will re-offend in the same conversation is 53\%, while the probability of another user using a personal attack at some point subsequently is 64\%.

\subsection{Individual user rebuttal levels}
\dsname{} contains utterances by 734 individuals. For users with more than 1 utterance (535 users), the median difference between the highest and lowest rebuttal level employed is 4, indicating that speakers utilise a range of strategies. 167 users were found to only use levels 4 and higher whereas 18 used only levels 3 and below. 

Mirroring \citep{meltzoff2002imitative} refers to a social phenomenon where speakers reflect the behaviour of others in a conversation. To identify whether this occurs in our data, we identify users who participated in more than one dispute (66 users). To determine whether a user $u$ is mirroring the rebuttal levels of others in a dispute $c$, we calculate:
\begin{align}
    m_{u,c} = \frac{\bar{r}_u - \bar{r}_{u,c}}{\bar{r}_u - \bar{r}_c},
\end{align}
where $\bar{r}_u$ represents a user's mean rebuttal level overall, $\bar{r}_{u,c}$ denotes the user's mean in dispute $c$, and $\bar{r}_{c}$ is dispute $c$'s mean rebuttal level excluding contributions from user $u$. $m_{u,c}$ represents the extent to which an individual's deviation from their own mean in a conversation is explained by the setting, and takes on a positive value if the changes are in the same direction. In our data, a positive $m_{u,c}$ is observed in 57\% of cases, indicating that mirroring does take place. 

\section{Modelling disagreements}\label{sec:methods}

\subsection{Multilabel classification} Predicting the dispute tactics used in an utterance is a multilabel classification task, meaning that for each utterance, any subset of $L$ classes (referred to as a \textit{labelset}) can be assigned as label. A vector representation for a labelset can be constructed as follows: $y_k = [y_{k1}, y_{k2}, ..., y_{kl}]$ where $y_{kj}=1$ $\iff$ the $j$-th of $L$ labels is relevant to the $k$-th example $x_k$ and 0 otherwise.

A simple method for multilabel modelling is \textbf{binary relevance (BR) classification}, under which a set of $L$ binary classifiers are trained to each predict independently whether a label applies or not \citep{boutell2004learning}. Another variation of the problem, referred to as the \textbf{label powerset (LP) method} \citep{tsoumakas2010random}, frames it as a multiclass classification problem over the powerset of all possible label combinations ($2^L$ combinations). Both of these approaches have been found to perform well in practice (see eg. \citealp{ferreira-vlachos-2019-incorporating}). 

With the advent of deep learning, \citet{nam2014large} proposed an alternative solution whereby the labelset vectors are directly predicted by a neural network model. This is achieved by modifying the output layer of a traditional multiclass neural network such that the SoftMax activation is replaced by a sigmoid function over every label in the output vector, and training with crossentropy loss. Since the sigmoid is applied to every output independently to determine its relevance to an utterance, we refer to this as a BR method, however, unlike other BR methods, the model is able to capture dependencies between classes.

\subsection{Metrics}\label{sec:metrics}
Evaluating multilabel models requires some consideration. Simple accuracy, also referred to as \textbf{exact match ratio (EMR)} in this setting, calculates the fraction of samples for which the full labelset are predicted correctly \citep{sorower2010literature}. This metric can be overly harsh, as it assigns no credit to partially correct predictions. A popular alternative is the \textbf{Hamming loss}, which measures the fraction of incorrectly assigned labels (or alternatively; the Hamming score measures the proportion of labels predicted correctly). For the $k$-th example with a predicted label vector $\hat{y_{k}}$, this would be:
\begin{align}
    Hamming(y_k,\hat{y_k}) = \frac{1}{L}\sum_{j=1}^{L}I[y_{kj}=\hat{y_{kj}}].
\end{align}
This metric can be overly generous in cases where the label assignment matrix is sparse, as is the case with our data; 75\% are assigned a single label out of 18 classes. The \textbf{Jaccard score} examines only the proportion of correctly predicted positive labels out of the potential positive set, and is seen as a midpoint between Hamming and EMR as two extremes \citep{park2018blended}:
\begin{align}
    Jaccard(y_k,\hat{y_k}) = \frac{|y_k \cap \hat{y_k}|}{|y_k \cup \hat{y_k}|}.
\end{align}
In our evaluation, we report all three metrics as they capture different perspectives, but prioritise the Jaccard score.  

\section{Experimental setting}

\paragraph{Truncated LP} As mentioned in Sec~\ref{sec:methods}, the LP method is a popular modelling choice for multilabel classification. In our case, a large number of labels are being considered relative to the number of samples, thus we do not use the full set of all label combinations, but instead consider only the 20 most commonly applied label sets in the training set. This method provides a coverage of 85\% of samples in the dataset. For samples whose labelsets are not in this category, we find the largest subset of their labels which does fall in the top 20 labelsets. If none of their labels qualify, which is the case for 175 utterances, we ignore these samples during training, but keep them in the test set. To allow for comparison between model types, we cast the predictions made by the LP model back to the multilabel setting for evaluation.

\paragraph{Incorporating conversation context}
While labels are assigned at the utterance level, some of the classes we aim to predict require knowledge of the preceding conversation context; for instance, ``repeated argument'' and ``refutation'' both occur in relation to another utterance. We experiment with both the context-agnostic method (to provide a baseline) and with models that incorporate the preceding context.

\paragraph{Predicting ordinality}
The abovementioned models do not incorporate knowledge of the preferential ordering of the rebuttal tactics. To include this signal, we add an auxiliary task which predicts the direction of the rebuttal level of the current utterance relative to the previous level, if a rebuttal tactic is used. Using Fig~\ref{fig:dh_sample} as an example, for utterance 2 the model would need to predict both level 4 (main task) and a downward transition (auxiliary task) relative to utterance 1. Further labels are added for upwards and same level transitions, as well as a separate class for coordination strategies which have no ordering. For the first utterance, level 3 is used as the reference level. If an utterance has multiple rebuttal labels, as indicated by the vertical lines in the figure, the maximum value was used as the reference point.  

\paragraph{Text encoders}\label{sec:models}
The following models are evaluated in our experiments: 
\begin{itemize}[noitemsep,nolistsep]
    \item \textbf{BoW}: A bag-of-words encoding of an utterance is processed by a two-layer multilayer perceptron. For the context-aware version, the preceding utterances are combined and encoded in the same manner, and the two vectors are concatenated as input to the model.
    \item \textbf{LSTM}: A recurrent neural network with long short-term memory \citep{hochreiter1997long} is used to process word embeddings (GloVe, \citealp{pennington-etal-2014-glove}). To incorporate conversation context, a hierarchical attention network (HAN; \citealp{yang2016hierarchical}) is used, which incorporates dialogue structure by encoding word embeddings  using an LSTM layer followed by an attention mechanism to build up utterance embeddings. Utterance embeddings are similarly combined to form a context embedding. 
    \item \textbf{BERT}: Two fully-connected layers are added to a pretrained transformer-based language model \citep{devlin2018bert} and finetuned for our classification task. We use the \textsc{``bert-base-cased"} model from \textsc{huggingface}. For the context-aware model, the preceding utterances are concatenated and encoded with a longformer model (\citealp{beltagy2020longformer}; \textsc{``allenai/longformer-base-4096"}). 
\end{itemize}

\paragraph{Implementation} We split the data into train, test and validation sets with a ratio of 70-20-10, and employ early stopping based on the validation loss. In the case of LP classification, the label with the largest score can be considered as the predicted class. In the BR setting a threshold must be set to determine whether a given label should be assigned. This is calibrated using a development set. All models are implemented in Keras. All models use dropout \citep{srivastava2014dropout} with $p=0.5$ and the Adam optimiser \citep{kingma2014adam} with learning rate 0.001; except for the transformer-based models which use a learning rate of 2e-5. 

\section{Results}\label{sec:results}
\begin{table}[t]
    \footnotesize
    \renewcommand{\arraystretch}{1.1}
    \centering
    \begin{tabular}{|l||r|r|r|}
        \hline
        \textbf{Model} & \textbf{Jaccard$\uparrow$} & \textbf{Hamming$\downarrow$} & \textbf{EMR$\uparrow$}\\
        \hline
        BoW BR &	0.2445&	0.1242	&0.105\\
        BoW LP & 0.3212	&0.096&0.2633\\
        BoW-CX BR &	0.3183&	0.1159	&0.1476\\
        BoW-CX LP &0.3218&	0.097&0.2722\\
        \hline
        LSTM BR&0.3263&0.1034&0.2259\\
        LSTM LP&0.2808&	0.1027	&0.2348\\
        LSTM-CX BR &0.332&	0.111&0.1459\\
        LSTM-CX LP &0.3164&0.0978&0.2669\\
        \hline
        BERT BR&0.3317&0.0875&0.2704\\	
        BERT LP&0.3339&0.0934&0.2687\\	
        BERT-CX BR&0.3451&	0.0875	&0.2705 \\
        BERT-CX LP &0.3505&0.0915&0.2882\\
        \hline
        BERT-CX-MT LP & \textbf{0.3858} & \textbf{0.0874} & \textbf{0.3344}\\
        \hline
        \end{tabular}
    \caption{Results for the binary relevance (BR) and truncated label powerset (LP) approaches, including conversation context (CX) and multitask modelling (MT). The best performance on each metric is shown in bold.}
    \label{tab:results}
\end{table}
Our experimental results are shown in Table \ref{tab:results}. We report results for the models described in Sec~\ref{sec:models}, using both context-agnostic and context-aware settings, with binary relevance and the truncated LP formulations. As per Sec~\ref{sec:metrics}, we report the Jaccard score, the Hamming loss and the EMR, but we prioritise the Jaccard score.

As expected, the LSTM and BERT models tend to outperform the BoW models. Adding conversation context improves performance in all models on the Jaccard score. The best performing model on the Jaccard metric is the BERT model with context.

The truncated LP method achieves better results for the majority of models and metrics compared to the binary relevance formulation, despite truncating the label powerset and ignoring 15\% of the training data. This can be attributed to the fact that there are certain classes with stronger co-occurrence relations (e.g.\  personal attacks with higher level arguments) which benefits the LP method. 

To gain a better understanding of model performance, we calculate the proportion of the test set with at least one label correctly predicted, yielding 0.395 on the best model (an increase of 0.11 over the EMR). The label most frequently correctly predicted is \textit{coordinating edits} (111 of 137 cases), which is also the most common label in the training set. The next most correctly predicted label, proportionally, is \textit{contextualisation} (75\%, or 24 of 32 cases), despite not being a commonly used label. This is likely due to the additional positional information available to the model, since this label is often applied to the first utterance in a conversation. On the other hand, \textit{refutation} and \textit{refuting the central point} are never correctly predicted (out of 44 cases), with \textit{counterargument} often mistakenly predicted instead. This is likely because of similarities between the classes and the latter being the second most common label in the training set.

We build on the best performing model to evaluate the effect of including the ordinality prediction auxiliary task. Using this model, we obtain a statistically significant improvement ($P=0.03$, using the permutation test on the Jaccard score) over the best model that is not aware of the ordering of the rebuttal tactics, providing further evidence of the usefulness of the rebuttal hierarchy. We also train models using the median and minimum rebuttal levels of each utterance, and find that while these do provide an increase in the Jaccard score, the difference is not significant ($P=0.32$ and $P=0.13$).   

\section{Predicting escalation}
Based on the findings of Sec~\ref{sec:analysis}, we believe that the dispute tactic annotations provided in the \dsname{} dataset can provide useful additional learning signals for the task of predicting escalation, as formulated by \citet{de-kock-vlachos-2021-beg}. We therefore do multitask training with escalation as the main task and tactics as the auxilliary task, such that the features that are predictive of dispute tactics are incorporated in the escalation predictions. 

\citet{de-kock-vlachos-2021-beg} found that a HAN network achieved the best results on this task; thus, we reproduce their experiment and achieve a PR-AUC score of 0.40. To add the dispute tactic classification as an auxiliary task, we modify the context-aware LSTM model, which uses a HAN model to encode the preceding context to an utterance (as described in Sec~\ref{sec:models}). To predict escalation for a given conversation, this HAN model is used to obtain a conversation embedding. Two further fully connected layers are added to allow for feature specialisation for the escalation prediction task. Using this model, a PR-AUC score of 0.487 is obtained, indicating that knowledge of these dispute tactics is useful for tasks beyond classifying the tactics employed.

\section{Conclusion}
In this work, we introduced a framework and dataset for dispute tactics, consisting of rebuttal and coordination strategies. We used these annotations to analyse how different tactics are used in disagreements and by different individuals, providing insights on the context in which personal attacks occur and the extent to which users alter their own tactics to mirror the level of rebuttal used in a conversation. We further develop multilabel models for classifying the dispute tactics used in an utterance, experimenting with both binary relevance and label powerset methods. We further illustrate that knowledge of these tactics increases accuracy on the task of predicting escalation, indicating that the framework and dataset can be of use to researchers working on other aspects of disagreements online.

\section*{Limitations}
Conversations on Wikipedia have been recognised for being goal-oriented \citep{niculae-danescu-niculescu-mizil-2016-conversational}, constructive \citep{de-kock-vlachos-2021-beg} and exhibiting low levels of toxicity \citep{wulczyn2017machina}. Learnings and models from Talk pages may therefore not transfer well to other domains.

Another limitation of this work is the size of the dataset, which is in part due to the difficulty of the task. Future work may look into creating larger datasets using this framework; however, as we have illustrated, the annotations can provide useful additional training signals for other conversation-based tasks such as predicting escalation. 

The dataset was annotated by only one annotator, which may introduce biases. For instance, cultural norms may influence what is considered to be a personal attack. To ameliorate this issue, an experienced annotator was consulted in the development of the framework. Three rounds of pilot annotation and discussion was carried out between the authors and the annotator, with increasing agreement scores.

As a factor of annotation ease, we only work with English data despite conversation data being available for many languages on the platform. Our insights, including the preferential rebuttal ordering, might not hold for other languages and contexts.

\section*{Acknowledgements}
Christine de Kock is supported by scholarships from Huawei and the Oppenheimer Memorial Trust. Andreas Vlachos and Tom Stafford are supported by the EPSRC grant Opening Up Minds (linked grants EP/T023414/1 and 
EP/T023554/1).  The
authors would like to acknowledge the work of Diane Nicholls in the annotation and the support of the Isaac Newton Trust and the Cambridge University Press.

\bibliography{custom}
\bibliographystyle{acl_natbib}

\appendix

\section{Graham's hierarchy of disagreement}\label{app:dh}
\begin{figure*}
    \centering
    \includegraphics[width=0.8\textwidth]{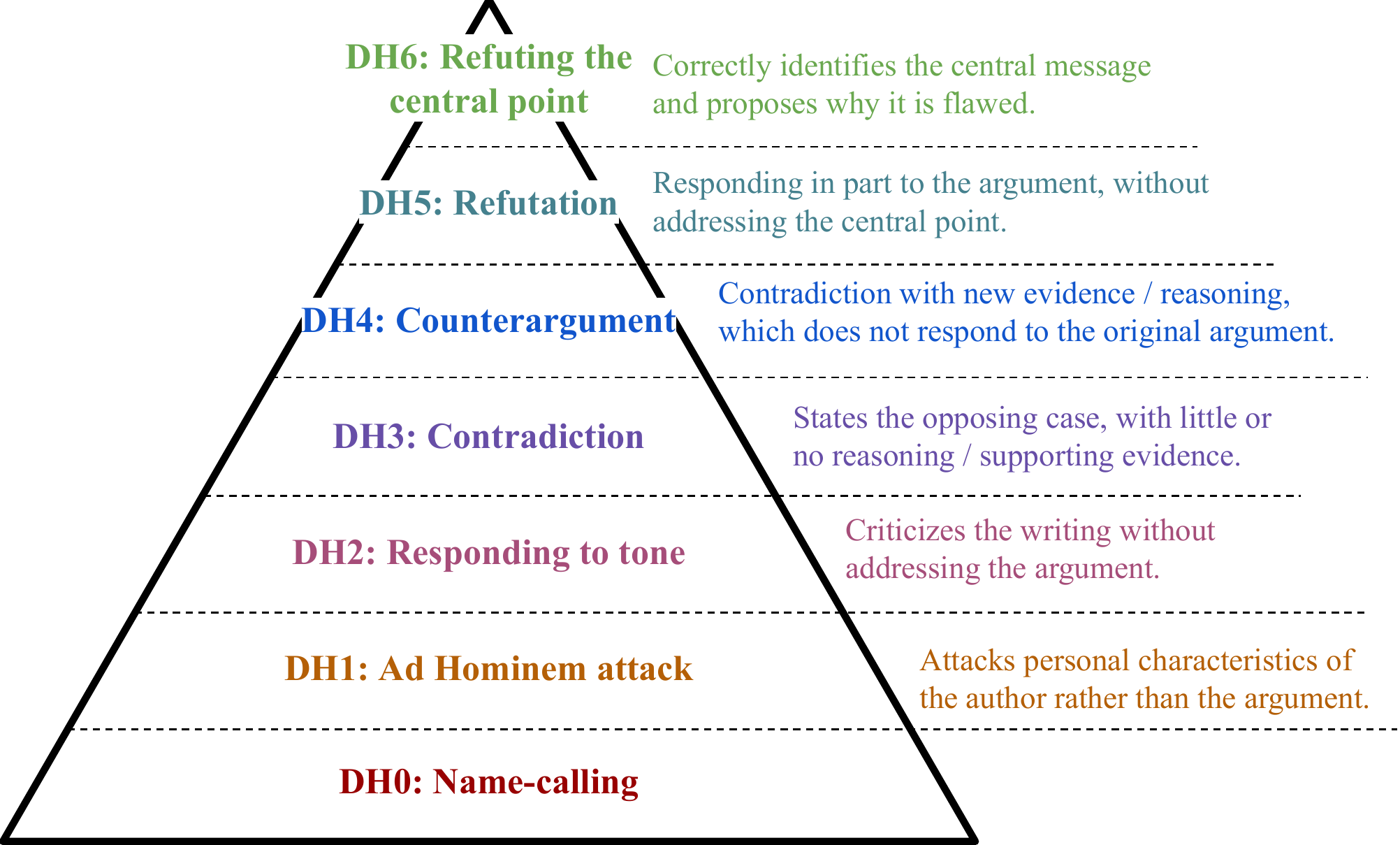}
    \caption{Illustration of Graham's Hierarchy of disagreement.}
    \label{fig:grahams_hierarchy}
\end{figure*}

An illustration of Graham's hierarchy of disagreement is shown in Figure \ref{fig:grahams_hierarchy}.

\section{Rebuttal tactics}\label{app:dh_schema}
The expanded disagreement hierarchy of rebuttal tactics and their definitions, as used in our annotations, are shown in Table \ref{tab:dh_schema}.

\begin{table*}[]
    \centering
    \begin{tabular}{|l|l|p{9cm}|}
    \hline
        \textbf{Level} & \textbf{Name} & \textbf{Description} \\\hline
        0& Name calling/hostility& Direct insults, or use of an equally hostile tone or language.\\\hline
        1& Ad hominem/ad argument& An attack to the person, often used to attempt to discredit an opponent (eg.\ ``I know better than you; you do not have a physics degree''). We extended this class to include ``ad argument'', where someone insults another's argument (eg.\ ``that is ridiculous'') without responding to its content.\\\hline
        2& \textcolor{red}{Attempted derailing/off-topic}& This category was added to address comments which are unrelated to the current line of discussion and fail to further the argument, while still being argumentative. This category is also represented in the Talk page taxonomy of \citet{viegas2007talk}. It was assigned a lower level as it can be detrimental to the argument, by taking focus away from the main topic. \\\hline
        3& Policing the discussion& \citet{graham} referred to this as “responding to tone” with the description “responses to the writing, rather than the writer. The lowest form of these is to disagree with the author's tone”. By expanding the definition to “policing the discussion”, the idea is to include people saying “You’ve said that before”, telling people to “calm down”, correcting spelling errors, or citing discussion policy (ie. “no personal attacks”). It ignores the argument’s content. \\\hline
        4& Stating your stance& \citet{graham} referred to this as ``contradiction'', with the description ``to state the opposing case, with little or no supporting evidence''. We renamed this class to include editors joining the conversation and just voicing their stance or their agreement with another editor.\\\hline
        4& \textcolor{red}{Repeated argument}& Level added to describe re-stating an argument used before, potentially using different words. This level was chosen as, similar to Level 4: Stating your stance, it does engage with the argument but does not further the discussion. \\\hline
        5& Counterargument& Described as ``contradiction plus new reasoning and/or evidence'', which does not directly address the opponent's argument.\\\hline
        6& Refutation& Directly responding to the argument and explaining why it is mistaken, using new evidence or reasoning.\\\hline
        7& Refuting the central point& \citet{graham} notes that ``Truly refuting something requires one to refute its central point, or at least one of them.'' Unfortunately, the central point can be quite subjective and difficult to recognise for non-experts, and may change throughout a conversation (as described in Sec~\ref{sec:pilot}). As such, we use this category as a prime example of a ``good'' refutation, which is of course still subjective and may be rolled up into Level 6. \\
        \hline
    \end{tabular}
    \caption{Rebuttal levels and descriptions. Levels added by us are indicated in red.}
    \label{tab:dh_schema}
\end{table*}

\section{Coordination tactics schema}\label{app:ndh_schema}
The coordination tactics and their definitions are shown in Table \ref{tab:nondh_schema}.
\begin{table*}[]
    \centering
    \begin{tabular}{|l|p{11cm}|}
        \hline
        \textbf{Name} & \textbf{Description}\\
        \hline
        Bailing out&An indication that an editor is giving up on a conversation and will no longer engage.\\\hline
        Contextualisation&In the first utterance, an editor ``sets the stage" by describing which aspect of the article they are challenging. This does not directly disagree with anyone, and is therefore a non-disagreement move.\\\hline
        Asking questions&Seeking to understand another editor's opinion better. This does not include rhetorical questions, which are generally disagreement moves.\\\hline
        Providing clarification&Answering questions or providing information which seeks to create understanding, rather than only furthering a point.\\\hline
        Suggesting a compromise&An attempt to find a midway between one's own point and the opposer's. This is explicitly encouraged by Wikipedia.\\\hline
        Coordinating edits&Wikipedia Talk pages are primarily used to for goal-oriented discussions, to coordinate edits to a page. As part of disagreement threads, there is often also some discussion of these edits. This can signal that a compromise has been found.\\\hline
        Conceding / recanting&An explicit admission that an interlocutor is willing to relinquish their point.\\\hline
        I don't know&Admitting that one is uncertain. This signals that an editor is receptive to the idea that there are unknowns which may impact their argument.\\\hline
        Other&For utterances not covered by any other class, for instance social niceties.\\
        \hline
    \end{tabular}
    \caption{Coordination tactic labels}
    \label{tab:nondh_schema}
\end{table*}

\section{Agreement}\label{app:agreement}
Inter-annotator agreement scores and other statistics of the pilot annotation rounds are shown in Table \ref{tab:agreement}.

\begin{table}[h]
    \centering
    \begin{tabular}{|l||l|l|l||l|l|}
    \hline
    \textbf{\#} &\textbf{\# utt.}&\textbf{A1 avg} & \textbf{A2 avg}& \textbf{$\kappa$} &  $r$ \\
    \hline
      1 & 51&3.86&2.98 &  0.17 & 0.57\\
      2 & 76&4.26&4.16 & 0.5  & 0.63\\
      3 & 67&3.93&4.06 & 0.55  & 0.68\\
    \hline
    \end{tabular}
    \caption{Statistics of the three pilot annotation rounds, including the number of utterances, the average rebuttal level assigned per annotator, and the Cohen's Kappa and Pearson's $r$.}
    \label{tab:agreement}
\end{table}

\section{PMI}\label{app:pmi}
The PMI scores of the different labels with personal attacks are shown in Table \ref{tab:pmi}.

\begin{table}[]
    \centering
    \begin{tabular}{|l|l|}
    \hline
    \textbf{Label} & \textbf{PMI} \\
    \hline
    RH6: Refutation &0.12\\
    RH5: Counterargument &0.07\\
    RH3: Policing the discussion &0.00\\
    Bailing out &-0.09\\
    RH4: Stating your stance &-0.20\\
    RH2: Off-topic comments &-0.33\\
    RH7: Refuting the central point &-0.42\\
    Suggesting a compromise &-1.57\\
    Coordinating edits &-1.75\\
    Providing clarification &-2.37\\
    Conceding / recanting &-2.85\\
    Asking questions &-2.96\\
    \hline
    \end{tabular}
    \caption{PMI of different labels with personal attacks.}
    \label{tab:pmi}
\end{table}

\end{document}